\documentclass{article}

\usepackage[nonatbib, final]{tackling_climate_workshop_style}

\usepackage[utf8]{inputenc} 
\usepackage[T1]{fontenc}    
\usepackage{hyperref}       
\usepackage{url}            
\usepackage{booktabs}       
\usepackage{amsfonts}       
\usepackage{nicefrac}       
\usepackage{microtype}      

\usepackage{amsmath}        
\usepackage{graphicx}       
\usepackage{float}          

\usepackage[
    backend=biber,
    style=numeric-comp,
    sorting=none,
    giveninits=true,
    maxbibnames=30,
    minbibnames=1,
    doi=true,
    url=false,
    isbn=false,
    eprint=false
]{biblatex}

\DeclareNameAlias{default}{family-given}
\DeclareNameAlias{sortname}{family-given}
\DeclareNameAlias{author}{family-given}
\renewbibmacro{in:}{}
\AtEveryBibitem{\clearfield{date}\clearfield{month}}

\DeclareFieldFormat{cite}{#1}

\title{Ranked Right, Sized Wrong: Ensemble Spread Fails Silently at Streamflow Extremes}

\author{%
  Justin Gallant\\
  University of British Columbia\\
  Vancouver, BC, Canada\\
  \texttt{gallant.justin.p@gmail.com}\\
  \And
  Tyler Wilson\\
  University of British Columbia\\
  Vancouver, BC, Canada\\
  \texttt{twilson@eoas.ubc.ca}
}

\begin{document}
 
\maketitle

\begin{abstract}
Deep learning models, particularly Long Short-Term Memory networks, are increasingly used for operational streamflow forecasting, with ensemble spread providing a signal of predictive uncertainty. As climate change increases the frequency of hydroclimatic extremes, these models will increasingly operate under flow conditions that are rare in their historical records. We therefore evaluate whether ensemble spread remains informative under increasing flows, examining both its magnitude relative to error and its ability to rank errors. Prediction error is approximately $2\times$ the corrected spread at ordinary flows and up to $4\times$ at the highest flows, with the discrepancy increasing with flow (log-log slope $1.30$, 95\% CI: $[1.20,1.41]$). Under-dispersion is strongest in observed extreme flows beyond the training record, reaching an $18\times$ discrepancy. This is pervasive and not explainable by conditional mean bias. Area Under the Sparsification Error, a standard distribution-free metric of uncertainty ranking, describing how well a model knows when it is uncertain, indicates near-oracle ranking for ensemble spread ($0.071$), yet ordering by predicted discharge alone nearly matches it. Informative uncertainty ranking can coexist with severe under-dispersion. In deployment, error ranking cannot be verified, yet the spread is available, and it is relied on as the uncertainty signal under increasingly extreme conditions, while systematically understating risk, most severely for the extreme events that matter most.
\end{abstract}

\section{Introduction}
Climate change is simultaneously increasing both the importance and difficulty of reliable hydrological prediction. As hydroclimatic extremes become more frequent under climate change \cite{seneviratneWeatherClimateExtreme2021}, forecasting systems will increasingly operate under conditions poorly represented in their historical record. Long Short-Term Memory (LSTM) networks have emerged as a leading approach for large-sample streamflow prediction, demonstrating high predictive skill by learning complex, nonlinear relationships directly from meteorological data \cite{kratzertRainfallRunoffModelling2018, leesBenchmarkingDatadrivenRainfall2021, maiGreatLakesRunoff2022, gauchRainfallRunoffPrediction2021} and are increasingly being adopted in operational forecasting applications \cite{huntUsingLongShortterm2022, kimPredictionFloodLevel2025}. However, as these models begin to support consequential climate adaptation decisions, evaluating mean predictive performance alone is no longer sufficient. A point prediction, no matter how skilled, is silent about how trustworthy it is, particularly when conditions depart from the model's historical experience.

Ensembles provide a practical approach for improving mean prediction while providing a signal of predictive uncertainty, and have been increasingly adopted in hydrological deep-learning (DL) applications \cite{willardMachineLearningEnsembles2025, chaudharyFloodUncertaintyEstimation2022}. While richer ensembling strategies can improve accuracy and probabilistic skill \cite{willardMachineLearningEnsembles2025}, standard practice in streamflow prediction often relies on random-initialization ensembles: multiple models sharing identical architectures and data, differing in their random initialization and data shuffling during optimization. The ensemble mean provides a point prediction, while the ensemble spread (between-member variance) is commonly interpreted as a signal of epistemic uncertainty: the reducible, model-side uncertainty associated with what the fit has not yet constrained \cite{lakshminarayananSimpleScalablePredictive2017, hullermeierAleatoricEpistemicUncertainty2021}. This is distinct from aleatoric uncertainty: the kind inherent to the observations, which is treated separately using methods designed to characterize irreducible variability \cite{klotzUncertaintyEstimationDeep2022, kraftModelingUncertaintyEngression2026}. \looseness=-1

The spread of a random-initialization ensemble is not guaranteed to reflect the magnitude of predictive error, even within the training regime, unlike deep ensembles trained with proper scoring rules \cite{lakshminarayananSimpleScalablePredictive2017}. Predictive error is expected to increase as inputs move into regions poorly represented by the training data \cite{frameDeepLearningRainfall2022}. We therefore expect spread to increase with error while remaining informative about the ordering of predictions by error \cite{ovadiaCanYouTrust2019}. As flows become increasingly extreme, the spread should therefore both increase as errors grow and remain informative about which predictions are most likely to be erroneous.\looseness=-1

In hydrological DL tasks, ensemble means are frequently used for prediction while raw ensemble spread is commonly reported as an uncertainty signal \cite{leesBenchmarkingDatadrivenRainfall2021, huntUsingLongShortterm2022, basteUnveilingLimitsDeep2025}. Existing work has examined ensemble calibration under in-distribution conditions \cite{chaudharyFloodUncertaintyEstimation2022} and developed approaches for uncertainty estimation under extrapolation or climate shift \cite{liuUncertaintyQuantificationMachine2023}. However, less is known about whether raw ensemble spread remains informative as flows become increasingly underrepresented in the training data. We therefore ask whether ensemble spread, the uncertainty signal available at deployment, remains reliable when realized error is unavailable. We evaluate reliability along two complementary dimensions: whether spread is appropriately scaled to error and whether it correctly ranks predictions by relative error, particularly as predictions move beyond the training distribution.\looseness=-1

\section{Methods}
We adopt a regional LSTM rainfall--runoff model developed for 269 unregulated catchments across Western Canada as a representative baseline model for evaluating uncertainty under hydroclimatic extremes (Sect.~\ref{sec:lstm_model}). The model uses daily precipitation and temperature with static catchment attributes (Table~\ref{tab:datasets}) and a one-year context window. It achieves a median test Nash--Sutcliffe Efficiency (NSE), a standard measure of streamflow prediction skill, of $0.75$, comparable to established regional rainfall--runoff benchmarks \cite{kratzertRainfallRunoffModelling2018, kratzertLearningUniversalRegional2019, andersonEvaluationInterpretationConvolutional2022}. Data are separated chronologically into training (1980--2005), validation (2006--2012), and testing (2013--2022) periods. Models are trained and selected using the training and validation periods and then evaluated without further adjustment on the held-out test period. The training period defines the fixed reference flow distribution used for all thresholds and exceedance levels.\looseness=-1

We construct a random-initialization ensemble of $M=10$ independently trained instances of the baseline model. For each test-set catchment-day, the ensemble mean ($\hat{\mu}$) provides the point prediction and the between-member variance ($\hat{\sigma}^2_e$) defines the ensemble spread used as the uncertainty signal. We evaluate two distinct properties of this signal: whether its dispersion appropriately reflects realized prediction error, and whether it correctly ranks predictions by realized error.\looseness=-1

To quantify residual dispersion relative to ensemble spread, we define the under-dispersion factor (UDF) for a stratum $\mathcal{S}$ as
\begin{equation}\label{eq:under-dispersion}
\text{UDF} \;=\; \frac{\mathrm{SD}(r\mid  \mathcal{S})}{\sqrt{\bar{\hat\sigma}^2_e}\sqrt{(M+1)/M}},
\end{equation}
where $r=y-\hat{\mu}$ (Sect.~\ref{sec:under-dispersion_factor}). Residual dispersion is calculated after removing the stratum mean to separate dispersion from conditional bias. The correction factor gives the null a value of $1$ under the exchangeability assumptions described in Sect.~\ref{sec:ensemble_correction}. Observation error and variability shared across ensemble members contribute to residual dispersion but are not represented in ensemble spread (Sect. \ref{sec:peak_bias}), meaning the null value of $1$ is a baseline. We focus on how the UDF changes with flow rather than interpreting its absolute magnitude as purely model-side uncertainty. We focus on the spread as it is the signal available at deployment. \looseness=-1

We examine ensemble spread across the observed flow distribution by stratifying catchment-days according to within-catchment observed-flow ($y$) percentiles and pooling corresponding deciles across catchments. Because observed flow is correlated with the residual, this conditioning can introduce bias. We therefore repeat the analysis using the ensemble mean prediction ($\hat\mu$) as the conditioning variable (Sect.~\ref{sec:peak_bias}). Conditional bias is reported as the mean residual within each stratum. To assess the ensemble spread scaling with realized error, we regress log residual standard deviation (SD) against log ensemble spread across deciles $10$--$90$. A slope of $1$ indicates proportional scaling, while slopes above $1$ indicate that error grows faster than the spread (Sect. \ref{sec:recalibration}). The top decile is held out to test extrapolation of the fitted relationship. Confidence intervals are obtained by resampling catchments with replacement ($B{=}1000$).\looseness=-1

To assess how dispersion changes toward increasingly rare high-flow conditions, we define thresholds relative to each catchment's observed 1980--2005 flow distribution. Test-period catchment-days are grouped by exceedance above these fixed training-period thresholds, including flows exceeding the historical maximum. This provides a basis for evaluating changes in dispersion across the upper tail.

Finally, we evaluate whether ensemble spread correctly ranks predictions by realized error. Since an ensemble of point predictors does not define a full predictive density, we use the distribution-free Area Under the Sparsification Error (AUSE) \cite{ilgUncertaintyEstimatesMultihypotheses2018} rather than likelihood-based scores. Catchment-days are progressively removed in descending order of the ranking variable, and Root Mean Squared Error (RMSE) of the retained predictions is compared with oracle (perfect ordering). Random ordering and ensemble mean ($\hat\mu$) are compared with ensemble spread. Lower AUSE indicates better ranking of high-error predictions. AUSE evaluates the ranking induced, not whether the magnitude is appropriate. \looseness=-1

\section{Results and Discussion}
\begin{figure}[!ht]
    \centering
    \includegraphics[width=1\linewidth]{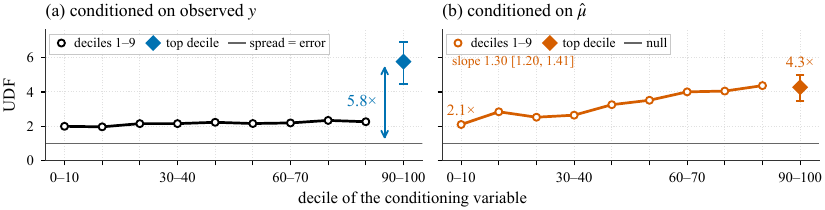}
    \caption{UDF by decile of the conditioning variable. \textbf{(a)} Deciles of observed discharge ($y$). \textbf{(b)} Deciles of the ensemble mean discharge ($\hat\mu$), with the fitted log-log scaling relationship and 95\% confidence interval. Error bars denote 95\% bootstrap intervals clustered by catchment. 
    }
    \label{fig:dispersion}
\end{figure} 

Figure~\ref{fig:dispersion} shows substantial under-dispersion of the ensemble across the flow distribution. Conditioned on observed flow (Fig.~\ref{fig:dispersion}a), the UDF increases from $2.1\times$ across the central 90\% of flows to $5.8\times$ in the highest decile.  Consistent with previously reported under-prediction of extreme streamflow by LSTMs \cite{frameDeepLearningRainfall2022}, the highest observed-flow decile also exhibits a $+1.304$~mm day$^{-1}$ conditional bias, compared with approximately ${-}0.15$~mm day$^{-1}$ across the remaining flow range. Conditioning on $\hat{\mu}$ substantially reduces this bias to approximately ${-}0.02$~mm day$^{-1}$, with $+0.25$~mm day$^{-1}$ in the highest decile. The under-dispersion persists with this correction in Fig.~\ref{fig:dispersion}b, with UDF increasing from $2.1\times$ to $4.3\times$ across ensemble-mean flow deciles and a superlinear slope identified by regressing log-UDF on log-flow. Thus, post-hoc recalibration should be flow-conditioned.\looseness=-1

\begin{figure}[!ht]
    \centering
    \includegraphics[width=1\linewidth]{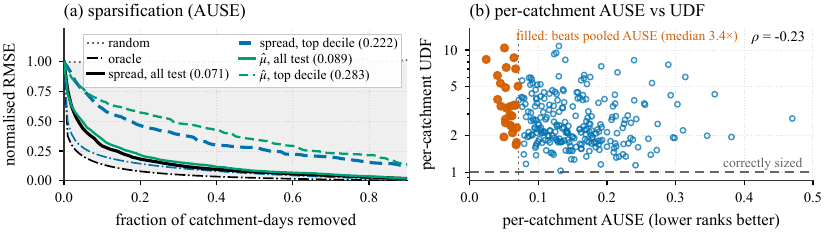}
    \caption{\textbf{(a)} Sparsification curves under ensemble spread and $\hat\mu$ for the full test set and top observed-flow decile, compared with oracle and random ordering. \textbf{(b)} AUSE versus UDF across catchments.\looseness=-1}
    \label{fig:shift}
\end{figure}

Despite this under-dispersion, ensemble spread performs well in a standard ranking check (Fig.~\ref{fig:shift}). The full-test-set sparsification curve approaches the oracle ordering, while performance degrades in the highest observed-flow decile, but remains substantially better than random ordering (Fig.~\ref{fig:shift}a). Ranking by $\hat\mu$ yields slightly higher AUSE values for the full test set and for the highest-flow decile (Fig.~\ref{fig:shift}a). The ensemble spread remains informative about relative prediction error, providing a signal of which predictions are likely to have larger errors. At the catchment level, under-dispersion is present in nearly every catchment despite this generally strong ranking performance (Fig.~\ref{fig:shift}b). One reason for this discrepancy is that ensemble spread reflects only disagreement between members: errors shared across members remain invisible, including systematic forcing error \cite{goswamiPerformanceEvaluationHighresolution2024, fatolahzadehgheysariReliabilityERA5ERA5Land2024}. Thus, an ensemble can remain informative about relative error while substantially underestimating its magnitude. \looseness=-2

\begin{table}[!ht]
    \centering
    \setlength{\tabcolsep}{4pt}
    \caption{UDF against each catchment's fixed 1980--2005 training flow distribution.
    }
    \label{tab:exceedance}
    \begin{tabular}{lrrrrrr}
    \toprule
    & $>$p90 & $>$p95 & $>$p99 & $>$p99.5 & $>$p99.9 & $>$max \\
    \midrule
    UDF & $5.65$ & $6.79$ & $10.40$ & $11.99$ & $14.41$ & $17.93$ \\
    observed / expected & $1.15$ & $1.26$ & $1.51$ & $1.64$ & $2.30$ & N/A \\
    catchment-days ($n$)   & $100{,}974$ & $55{,}212$ & $13{,}295$ & $7{,}185$ & $2{,}023$ & $603$ \\
    \bottomrule
    \end{tabular}
\end{table}

The UDF increases substantially toward the upper tail of the training-period flow distribution, rising from $5.65\times$ above the training-period 90th percentile to $17.93\times$ for flows exceeding the historical maximum (Table~\ref{tab:exceedance}). These extreme observations may also carry greater observational uncertainty because their rating curves require extrapolation \cite{morlotDynamicRatingCurve2014, mansanarezRapidStageDischargeRating2019} (Sect. \ref{sec:peak_bias}). Nevertheless, observed exceedances of the fixed 1980--2005 reference distribution become increasingly frequent during the test period, exposing the model to progressively rarer high-flow conditions, under which the ensemble's under-dispersion worsens dramatically. \looseness=-1

\section{Conclusion}
Ensembles of LSTMs are widely adopted for streamflow prediction in hydrological DL and perform generally well. We evaluated whether the spread of a field-standard random-initialization ensemble provides a reliable uncertainty signal by examining both its dispersion relative to realized prediction error and its ability to rank prediction error by uncertainty. The ensemble spread consistently underestimates the magnitude of prediction error, worsening at extreme flows. Meanwhile, the spread achieves near-oracle AUSE, remaining informative about the relative magnitude of prediction error, despite degradation at extremes and only modestly outperforming the ensemble mean. Thus, an ensemble can successfully rank predictions by error while substantially underestimating the magnitude of that error. \looseness=-1 

This is the trap: in deployment, there is no ground truth at the time of an event; the ensemble spread exists, but the streamflow residual does not. Under-dispersion is silent, and it worsens as flows become increasingly extreme and underrepresented in the training. As high flow events become more frequent, more decisions will rest on these models in precisely the regimes where their uncertainty is least trustworthy. The under-dispersion identified is pervasive, indicating that improving the mean performance alone is unlikely to resolve it. Ranking alone is insufficient for uncertainty evaluation; it should therefore assess both dispersion and ranking. Since error grows faster than spread, no constant post-hoc rescaling can correct this; the correction should therefore be conditioned on flow, such as using conditional post-processing or distributional heads \cite{lakshminarayananSimpleScalablePredictive2017, klotzUncertaintyEstimationDeep2022}, which we explore in ongoing work.

\begin{ack}
Funding supporting this study was provided by the Natural Sciences and Engineering Research Council of Canada (NSERC) through an Undergraduate Student Research Award (USRA) to JG and a Master's Scholarship to TW.

This research was supported in part through the computational resources and services provided by Advanced Research Computing at the University of British Columbia.
\end{ack}

\medskip

\begin{small}
\printbibliography[title={References}]

@article{andersonEvaluationInterpretationConvolutional2022,
  title = {Evaluation and Interpretation of Convolutional Long Short-Term Memory Networks for Regional Hydrological Modelling},
  author = {Anderson, Sam and Radi{\'c}, Valentina},
  year = 2022,
  month = feb,
  journal = {Hydrology and Earth System Sciences},
  volume = {26},
  number = {3},
  pages = {795--825},
  issn = {1607-7938},
  doi = {10.5194/hess-26-795-2022},
  url = {https://hess.copernicus.org/articles/26/795/2022/},
  copyright = {https://creativecommons.org/licenses/by/4.0/},
  langid = {english}
}

@article{basteUnveilingLimitsDeep2025,
  title = {Unveiling the Limits of Deep Learning Models in Hydrological Extrapolation Tasks},
  author = {Baste, Sanika and Klotz, Daniel and Acu{\~n}a Espinoza, Eduardo and Bardossy, Andras and Loritz, Ralf},
  year = 2025,
  month = nov,
  journal = {Hydrology and Earth System Sciences},
  volume = {29},
  number = {21},
  pages = {5871--5891},
  issn = {1607-7938},
  doi = {10.5194/hess-29-5871-2025},
  url = {https://hess.copernicus.org/articles/29/5871/2025/},
  copyright = {https://creativecommons.org/licenses/by/4.0/},
  langid = {english}
}

@article{chaudharyFloodUncertaintyEstimation2022,
  title = {Flood {{Uncertainty Estimation Using Deep Ensembles}}},
  author = {Chaudhary, Priyanka and Leit{\~a}o, Jo{\~a}o P. and Donauer, Tabea and D'Aronco, Stefano and Perraudin, Nathana{\"e}l and Obozinski, Guillaume and {Perez-Cruz}, Fernando and Schindler, Konrad and Wegner, Jan Dirk and Russo, Stefania},
  year = 2022,
  month = sep,
  journal = {Water},
  volume = {14},
  number = {19},
  pages = {2980},
  issn = {2073-4441},
  doi = {10.3390/w14192980},
  url = {https://www.mdpi.com/2073-4441/14/19/2980},
  langid = {english}
}

@misc{copernicusclimatechangeservicec3sERA5LandHourlyData2019,
  title = {{{ERA5-Land}} Hourly Data from 1950 to Present},
  author = {{Copernicus Climate Change Service (C3S)}},
  year = 2019,
  publisher = {Copernicus Climate Change Service (C3S) Climate Data Store (CDS)},
  doi = {10.24381/CDS.E2161BAC},
  url = {https://cds.climate.copernicus.eu/doi/10.24381/cds.e2161bac},
  howpublished = {Copernicus Climate Change Service (C3S) Climate Data Store (CDS) [data set]}
}

@misc{environmentandclimatechangecanadaNationalHydrometricNetwork2024,
  title = {National Hydrometric Network Basin Polygons},
  author = {{Environment and Climate Change Canada}},
  year = 2024,
  month = aug,
  url = {https://collaboration.cmc.ec.gc.ca/cmc/hydrometrics/www/HydrometricNetworkBasinPolygons/},
  howpublished = {Environment and Climate Change Canada [data set]},
  note = {(last access: 11 February 2026)}
}

@article{fatolahzadehgheysariReliabilityERA5ERA5Land2024,
  title = {Reliability of {{ERA5}} and {{ERA5-Land}} Reanalysis Data in the {{Canadian Prairies}}},
  author = {Fatolahzadeh Gheysari, Ali and Maghoul, Pooneh and Ojo, E. RoTimi and Shalaby, Ahmed},
  year = 2024,
  month = apr,
  journal = {Theoretical and Applied Climatology},
  volume = {155},
  number = {4},
  pages = {3087--3098},
  issn = {0177-798X, 1434-4483},
  doi = {10.1007/s00704-023-04771-z},
  url = {https://link.springer.com/10.1007/s00704-023-04771-z},
  langid = {english}
}

@article{fortinWhyShouldEnsemble2014,
  title = {Why {{Should Ensemble Spread Match}} the {{RMSE}} of the {{Ensemble Mean}}?},
  author = {Fortin, V. and Abaza, M. and Anctil, F. and Turcotte, R.},
  year = 2014,
  month = aug,
  journal = {Journal of Hydrometeorology},
  volume = {15},
  number = {4},
  pages = {1708--1713},
  issn = {1525-755X, 1525-7541},
  doi = {10.1175/JHM-D-14-0008.1},
  url = {http://journals.ametsoc.org/doi/10.1175/JHM-D-14-0008.1},
  langid = {english}
}

@article{frameDeepLearningRainfall2022,
  title = {Deep Learning Rainfall--Runoff Predictions of Extreme Events},
  author = {Frame, Jonathan M. and Kratzert, Frederik and Klotz, Daniel and Gauch, Martin and Shalev, Guy and Gilon, Oren and Qualls, Logan M. and Gupta, Hoshin V. and Nearing, Grey S.},
  year = 2022,
  month = jul,
  journal = {Hydrology and Earth System Sciences},
  volume = {26},
  number = {13},
  pages = {3377--3392},
  issn = {1607-7938},
  doi = {10.5194/hess-26-3377-2022},
  url = {https://hess.copernicus.org/articles/26/3377/2022/},
  copyright = {https://creativecommons.org/licenses/by/4.0/},
  langid = {english}
}

@misc{gallantLSTMEnsembleUnderdispersion2026,
  author = {Gallant, Justin and Wilson, Tyler},
  year = 2026,
  month = aug,
  doi = {10.5281/zenodo.23142989},
  copyright = {Creative Commons Attribution 4.0 International},
  title = {{{LSTM Ensemble Under-dispersion Code}} ({{Version}} 1.1.0)},
  howpublished = {Zenodo [code]}
}

@article{gauchRainfallRunoffPrediction2021,
  title = {Rainfall--Runoff Prediction at Multiple Timescales with a Single {{Long Short-Term Memory}} Network},
  author = {Gauch, Martin and Kratzert, Frederik and Klotz, Daniel and Nearing, Grey and Lin, Jimmy and Hochreiter, Sepp},
  year = 2021,
  month = apr,
  journal = {Hydrology and Earth System Sciences},
  volume = {25},
  number = {4},
  pages = {2045--2062},
  issn = {1607-7938},
  doi = {10.5194/hess-25-2045-2021},
  url = {https://hess.copernicus.org/articles/25/2045/2021/},
  copyright = {https://creativecommons.org/licenses/by/4.0/},
  langid = {english}
}

@article{goswamiPerformanceEvaluationHighresolution2024,
  title = {Performance {{Evaluation}} of {{High-resolution Reanalysis Datasets Over North-central British Columbia}}},
  author = {Goswami, Uttam P. and D{\'e}ry, Stephen J. and Fortin, Vincent},
  year = 2024,
  month = may,
  journal = {Atmosphere-Ocean},
  volume = {62},
  number = {3},
  pages = {222--242},
  issn = {0705-5900, 1480-9214},
  doi = {10.1080/07055900.2024.2308878},
  url = {https://www.tandfonline.com/doi/full/10.1080/07055900.2024.2308878},
  langid = {english}
}

@article{hullermeierAleatoricEpistemicUncertainty2021,
  title = {Aleatoric and Epistemic Uncertainty in Machine Learning: An Introduction to Concepts and Methods},
  shorttitle = {Aleatoric and Epistemic Uncertainty in Machine Learning},
  author = {H{\"u}llermeier, Eyke and Waegeman, Willem},
  year = 2021,
  month = mar,
  journal = {Machine Learning},
  volume = {110},
  number = {3},
  pages = {457--506},
  issn = {0885-6125, 1573-0565},
  doi = {10.1007/s10994-021-05946-3},
  url = {https://link.springer.com/10.1007/s10994-021-05946-3},
  langid = {english}
}

@article{huntUsingLongShortterm2022,
  title = {Using a Long Short-Term Memory ({{LSTM}}) Neural Network to Boost River Streamflow Forecasts over the Western {{United States}}},
  author = {Hunt, Kieran M. R. and Matthews, Gwyneth R. and Pappenberger, Florian and Prudhomme, Christel},
  year = 2022,
  month = nov,
  journal = {Hydrology and Earth System Sciences},
  volume = {26},
  number = {21},
  pages = {5449--5472},
  issn = {1607-7938},
  doi = {10.5194/hess-26-5449-2022},
  url = {https://hess.copernicus.org/articles/26/5449/2022/},
  copyright = {https://creativecommons.org/licenses/by/4.0/},
  langid = {english}
}

@inproceedings{ilgUncertaintyEstimatesMultihypotheses2018,
  title = {Uncertainty {{Estimates}} and {{Multi-hypotheses Networks}} for {{Optical Flow}}},
  booktitle = {Computer {{Vision}} -- {{ECCV}} 2018},
  author = {Ilg, Eddy and {\c C}i{\c c}ek, {\"O}zg{\"u}n and Galesso, Silvio and Klein, Aaron and Makansi, Osama and Hutter, Frank and Brox, Thomas},
  editor = {Ferrari, Vittorio and Hebert, Martial and Sminchisescu, Cristian and Weiss, Yair},
  year = 2018,
  month = {9},
  volume = {11211},
  pages = {677--693},
  publisher = {Springer International Publishing},
  address = {Munich, Germany},
  doi = {10.1007/978-3-030-01234-2_40},
  url = {https://link.springer.com/10.1007/978-3-030-01234-2_40},
  isbn = {978-3-030-01234-2},
  langid = {english}
}

@article{kimPredictionFloodLevel2025,
  title = {Prediction of {{Flood Level Using}} {{{\textsc{LSTM}}}} and {{Watershed Hydrological Data}}},
  shorttitle = {Prediction of {{Flood Level Using}}},
  author = {Kim, Hyun-il and Jang, Se-Dong and Choi, Hehun and Kim, Tae-Hyung and Kim, Byunghyun},
  year = 2025,
  month = dec,
  journal = {Journal of Flood Risk Management},
  volume = {18},
  number = {4},
  pages = {e70123},
  issn = {1753-318X, 1753-318X},
  doi = {10.1111/jfr3.70123},
  url = {https://onlinelibrary.wiley.com/doi/10.1111/jfr3.70123},
  langid = {english}
}

@article{klotzUncertaintyEstimationDeep2022,
  title = {Uncertainty Estimation with Deep Learning for Rainfall--Runoff Modeling},
  author = {Klotz, Daniel and Kratzert, Frederik and Gauch, Martin and Keefe Sampson, Alden and Brandstetter, Johannes and Klambauer, G{\"u}nter and Hochreiter, Sepp and Nearing, Grey},
  year = 2022,
  month = mar,
  journal = {Hydrology and Earth System Sciences},
  volume = {26},
  number = {6},
  pages = {1673--1693},
  issn = {1607-7938},
  doi = {10.5194/hess-26-1673-2022},
  url = {https://hess.copernicus.org/articles/26/1673/2022/},
  copyright = {https://creativecommons.org/licenses/by/4.0/},
  langid = {english}
}

@article{kraftModelingUncertaintyEngression2026,
  title = {Modeling {{Uncertainty With Engression}}: {{A Deep Generative Time}}-{{Series Approach}}},
  shorttitle = {Modeling {{Uncertainty With Engression}}},
  author = {Kraft, Basil and Stalder, Steven and Aeberhard, William H. and Ruiz, Nicol{\'a}s Harrington and Meinshausen, Nicolai and Shen, Xinwei and Gudmundsson, Lukas},
  year = 2026,
  month = jan,
  journal = {Geophysical Research Letters},
  volume = {53},
  number = {2},
  pages = {e2025GL120122},
  issn = {0094-8276, 1944-8007},
  doi = {10.1029/2025GL120122},
  url = {https://agupubs.onlinelibrary.wiley.com/doi/10.1029/2025GL120122},
  langid = {english}
}

@article{kratzertLearningUniversalRegional2019,
  title = {Towards Learning Universal, Regional, and Local Hydrological Behaviors via Machine Learning Applied to Large-Sample Datasets},
  author = {Kratzert, Frederik and Klotz, Daniel and Shalev, Guy and Klambauer, G{\"u}nter and Hochreiter, Sepp and Nearing, Grey},
  year = 2019,
  month = dec,
  journal = {Hydrology and Earth System Sciences},
  volume = {23},
  number = {12},
  pages = {5089--5110},
  issn = {1607-7938},
  doi = {10.5194/hess-23-5089-2019},
  url = {https://hess.copernicus.org/articles/23/5089/2019/},
  copyright = {https://creativecommons.org/licenses/by/4.0/},
  langid = {english}
}

@article{kratzertRainfallRunoffModelling2018,
  title = {Rainfall--Runoff Modelling Using {{Long Short-Term Memory}} ({{LSTM}}) Networks},
  author = {Kratzert, Frederik and Klotz, Daniel and Brenner, Claire and Schulz, Karsten and Herrnegger, Mathew},
  year = 2018,
  month = nov,
  journal = {Hydrology and Earth System Sciences},
  volume = {22},
  number = {11},
  pages = {6005--6022},
  issn = {1607-7938},
  doi = {10.5194/hess-22-6005-2018},
  url = {https://hess.copernicus.org/articles/22/6005/2018/},
  copyright = {https://creativecommons.org/licenses/by/4.0/},
  langid = {english}
}

@inproceedings{lakshminarayananSimpleScalablePredictive2017,
  title = {Simple and Scalable Predictive Uncertainty Estimation Using Deep Ensembles},
  booktitle = {Advances in {{Neural Information Processing Systems}} 30},
  author = {Lakshminarayanan, Balaji and Pritzel, Alexander and Blundell, Charles},
  editor = {Guyon, Isabelle and {von Luxburg}, Ulrike and Bengio, Samy and Wallach, Hanna M. and Fergus, Rob and Vishwanathan, S. V. N. and Garnett, Roman},
  year = 2017,
  month = {12},
  volume = {30},
  pages = {6402--6413},
  publisher = {Curran Associates, Inc.},
  address = {Long Beach, CA, USA},
  url = {https://proceedings.neurips.cc/paper/2017/hash/9ef2ed4b7fd2c810847ffa5fa85bce38-Abstract.html},
}

@article{leesBenchmarkingDatadrivenRainfall2021,
  title = {Benchmarking Data-Driven Rainfall--Runoff Models in {{Great Britain}}: A Comparison of Long Short-Term Memory ({{LSTM}})-Based Models with Four Lumped Conceptual Models},
  shorttitle = {Benchmarking Data-Driven Rainfall--Runoff Models in {{Great Britain}}},
  author = {Lees, Thomas and Buechel, Marcus and Anderson, Bailey and Slater, Louise and Reece, Steven and Coxon, Gemma and Dadson, Simon J.},
  year = 2021,
  month = oct,
  journal = {Hydrology and Earth System Sciences},
  volume = {25},
  number = {10},
  pages = {5517--5534},
  issn = {1607-7938},
  doi = {10.5194/hess-25-5517-2021},
  url = {https://hess.copernicus.org/articles/25/5517/2021/},
  copyright = {https://creativecommons.org/licenses/by/4.0/},
  langid = {english}
}

@article{liuUncertaintyQuantificationMachine2023,
  title = {Uncertainty Quantification of Machine Learning Models to Improve Streamflow Prediction under Changing Climate and Environmental Conditions},
  author = {Liu, Siyan and Lu, Dan and Painter, Scott L. and Griffiths, Natalie A. and Pierce, Eric M.},
  year = 2023,
  month = apr,
  journal = {Frontiers in Water},
  volume = {5},
  pages = {1150126},
  issn = {2624-9375},
  doi = {10.3389/frwa.2023.1150126},
  url = {https://www.frontiersin.org/articles/10.3389/frwa.2023.1150126/full}
}

@article{maiGreatLakesRunoff2022,
  title = {The {{Great Lakes Runoff Intercomparison Project Phase}} 4: The {{Great Lakes}} ({{GRIP-GL}})},
  shorttitle = {The {{Great Lakes Runoff Intercomparison Project Phase}} 4},
  author = {Mai, Juliane and Shen, Hongren and Tolson, Bryan A. and Gaborit, {\'E}tienne and Arsenault, Richard and Craig, James R. and Fortin, Vincent and Fry, Lauren M. and Gauch, Martin and Klotz, Daniel and Kratzert, Frederik and O'Brien, Nicole and Princz, Daniel G. and Rasiya Koya, Sinan and Roy, Tirthankar and Seglenieks, Frank and Shrestha, Narayan K. and Temgoua, Andr{\'e} G. T. and Vionnet, Vincent and Waddell, Jonathan W.},
  year = 2022,
  month = jul,
  journal = {Hydrology and Earth System Sciences},
  volume = {26},
  number = {13},
  pages = {3537--3572},
  issn = {1607-7938},
  doi = {10.5194/hess-26-3537-2022},
  url = {https://hess.copernicus.org/articles/26/3537/2022/},
  copyright = {https://creativecommons.org/licenses/by/4.0/},
  langid = {english}
}

@article{mansanarezRapidStageDischargeRating2019,
  title = {Rapid {{Stage}}-{{Discharge Rating Curve Assessment Using Hydraulic Modeling}} in an {{Uncertainty Framework}}},
  author = {Mansanarez, Valentin and Westerberg, Ida K. and Lam, Norris and Lyon, Steve W.},
  year = 2019,
  month = nov,
  journal = {Water Resources Research},
  volume = {55},
  number = {11},
  pages = {9765--9787},
  issn = {0043-1397, 1944-7973},
  doi = {10.1029/2018WR024176},
  url = {https://agupubs.onlinelibrary.wiley.com/doi/10.1029/2018WR024176},
  langid = {english}
}

@misc{mapzenTerrainTiles2026,
  title = {Terrain {{Tiles}}},
  shorttitle = {{{AWS Terrain Tiles}}},
  author = {{Mapzen}},
  year = 2026,
  publisher = {Amazon S3 (Open Data Registry)},
  url = {https://registry.opendata.aws/terrain-tiles/},
  archiveprefix = {Registry of Open Data on AWS},
  copyright = {Creative Commons license specified at: https://github.com/tilezen/joerd/blob/master/docs/attribution.md},
  howpublished = {Amazon S3 (Open Data Registry) [data set]},
  note = {(last access: 11 February 2026)}
}

@article{morlotDynamicRatingCurve2014,
  title = {Dynamic Rating Curve Assessment for Hydrometric Stations and Computation of the Associated Uncertainties: {{Quality}} and Station Management Indicators},
  shorttitle = {Dynamic Rating Curve Assessment for Hydrometric Stations and Computation of the Associated Uncertainties},
  author = {Morlot, Thomas and Perret, Christian and Favre, Anne-Catherine and Jalbert, Jonathan},
  year = 2014,
  month = sep,
  journal = {Journal of Hydrology},
  volume = {517},
  pages = {173--186},
  issn = {00221694},
  doi = {10.1016/j.jhydrol.2014.05.007},
  url = {https://linkinghub.elsevier.com/retrieve/pii/S0022169414003618},
  langid = {english}
}

@misc{munozsabaterERA5LandHourlyData2019,
  title = {{{ERA5-Land}} Hourly Data from 1950 to Present},
  author = {Mu{\~n}oz Sabater, J.},
  year = 2019,
  publisher = {Copernicus Climate Change Service (C3S) Climate Data Store (CDS)},
  doi = {10.24381/cds.e2161bac},
  url = {https://cds.climate.copernicus.eu/doi/10.24381/cds.e2161bac},
  howpublished = {Copernicus Climate Change Service (C3S) Climate Data Store (CDS) [data set]},
  note = {(last access: 23 February 2026)}
}

@article{nashRiverFlowForecasting1970,
  title = {River Flow Forecasting through Conceptual Models Part {{I}} --- {{A}} Discussion of Principles},
  author = {Nash, J.E. and Sutcliffe, J.V.},
  year = 1970,
  month = apr,
  journal = {Journal of Hydrology},
  volume = {10},
  number = {3},
  pages = {282--290},
  issn = {00221694},
  doi = {10.1016/0022-1694(70)90255-6},
  url = {https://linkinghub.elsevier.com/retrieve/pii/0022169470902556},
  copyright = {https://www.elsevier.com/tdm/userlicense/1.0/},
  langid = {english}
}

@inproceedings{ovadiaCanYouTrust2019,
  title = {Can {{You Trust Your Model}}'s {{Uncertainty}}? {{Evaluating Predictive Uncertainty Under Dataset Shift}}},
  booktitle = {Advances in {{Neural Information Processing Systems}} 32},
  author = {Ovadia, Yaniv and Fertig, Emily and Ren, Jie and Nado, Zachary and Sculley, D. and Nowozin, Sebastian and Dillon, Joshua and Lakshminarayanan, Balaji and Snoek, Jasper},
  year = 2019,
  month = {12},
  volume = {32},
  pages = {13969--13980},
  publisher = {Curran Associates, Inc.},
  address = {Vancouver, BC, Canada},
  url = {https://proceedings.neurips.cc/paper/2019/hash/8558cb408c1d76621371888657d2eb1d-Abstract.html}
}

@misc{rgiconsortiumRandolphGlacierInventory2017,
  author = {{RGI Consortium}},
  year = 2017,
  month = jul,
  publisher = {{NSIDC: National Snow and Ice Data Center}},
  address = {Boulder, Colorado USA},
  doi = {10.7265/N5-RGI-60},
  url = {http://www.glims.org/RGI/randolph60.html},
  langid = {english},
  title = {Randolph {{Glacier Inventory}} - {{A Dataset}} of {{Global Glacier Outlines}} ({{Version}} 6.0)},
  howpublished = {NSIDC: National Snow and Ice Data Center [data set]}
}

@article{schnorbusImpactsClimateChange2014,
  title = {Impacts of Climate Change in Three Hydrologic Regimes in {{British Columbia}}, {{Canada}}},
  shorttitle = {Impacts of Climate Change in Three Hydrologic Regimes in {{British Columbia}}, {{Canada}}},
  author = {Schnorbus, Markus and Werner, Arelia and Bennett, Katrina},
  year = 2014,
  month = jan,
  journal = {Hydrological Processes},
  volume = {28},
  number = {3},
  pages = {1170--1189},
  issn = {08856087},
  doi = {10.1002/hyp.9661},
  url = {https://onlinelibrary.wiley.com/doi/10.1002/hyp.9661},
  copyright = {http://doi.wiley.com/10.1002/tdm\_license\_1.1},
  langid = {english}
}

@incollection{seneviratneWeatherClimateExtreme2021,
  title = {Weather and Climate Extreme Events in a Changing Climate},
  booktitle = {Climate {{Change}} 2021 -- {{The Physical Science Basis Working Group I Contribution}} to the {{Sixth Assessment Report}} of the {{Intergovernmental Panel}} on {{Climate Change}}},
  author = {Seneviratne, S. I. and Zhang, X. and Adnan, M. and Badi, W. and Dereczynski, C. and Di Luca, A. and Ghosh, S. and Iskandar, I. and Kossin, J. and Lewis, S. and Otto, F. and Pinto, I. and Satoh, M. and {Vicente-Serrano}, S. M. and Wehner, M. and Zhou, B.},
  editor = {{Masson-Delmotte}, V. and Zhai, P. and Pirani, A. and Connors, S. L. and P{\'e}an, C. and Berger, S. and Caud, N. and Chen, Y. and Goldfarb, L. and Gomis, M. I. and Huang, M. and Leitzell, K. and Lonnoy, E. and Matthews, J. B. R. and Maycock, T. K. and Waterfield, T. and Yelek{\c c}i, O. and Yu, R. and Zhou, B.},
  year = 2021,
  edition = {1},
  pages = {1513--1766},
  publisher = {Cambridge University Press},
  doi = {10.1017/9781009157896.013},
  copyright = {https://www.cambridge.org/core/terms}
}

@misc{watersurveyofcanadaEnvironmentClimateChange2026,
  title = {Environment and {{Climate Change Canada Historical Hydrometric Data}} ({{HYDAT}})},
  author = {{Water Survey of Canada}},
  year = 2026,
  url = {https://wateroffice.ec.gc.ca/mainmenu/historical_data_index_e.html},
  howpublished = {Water Survey of Canada [data set]},
  note = {(last access: 11 February 2026)}
}

@article{whiteUnprecedentedPacificNorthwest2023,
  title = {The Unprecedented {{Pacific Northwest}} Heatwave of {{June}} 2021},
  author = {White, Rachel H. and Anderson, Sam and Booth, James F. and Braich, Ginni and Draeger, Christina and Fei, Cuiyi and Harley, Christopher D. G. and Henderson, Sarah B. and Jakob, Matthias and Lau, Carie-Ann and Mareshet Admasu, Lualawi and Narinesingh, Veeshan and Rodell, Christopher and Roocroft, Eliott and Weinberger, Kate R. and West, Greg},
  year = 2023,
  month = feb,
  journal = {Nature Communications},
  volume = {14},
  number = {1},
  pages = {727},
  issn = {2041-1723},
  doi = {10.1038/s41467-023-36289-3},
  url = {https://www.nature.com/articles/s41467-023-36289-3},
  langid = {english}
}

@article{willardMachineLearningEnsembles2025,
  title = {Machine {{Learning Ensembles Can Enhance Hydrologic Predictions}} and {{Uncertainty Quantification}}},
  author = {Willard, Jared D. and Varadharajan, Charuleka},
  year = 2025,
  month = sep,
  journal = {Journal of Geophysical Research: Machine Learning and Computation},
  volume = {2},
  number = {3},
  pages = {e2025JH000732},
  issn = {2993-5210, 2993-5210},
  doi = {10.1029/2025JH000732},
  url = {https://agupubs.onlinelibrary.wiley.com/doi/10.1029/2025JH000732},
  langid = {english}
}

@inproceedings{wimmerQuantifyingAleatoricEpistemic31,
  title = {Quantifying {{Aleatoric}} and {{Epistemic Uncertainty}} in {{Machine Learning}}: {{Are Conditional Entropy}} and {{Mutual Information Appropriate Measures}}?},
  shorttitle = {Quantifying {{Uncertainty}} in {{ML}}},
  booktitle = {Proceedings of the {{Thirty-Ninth Conference}} on {{Uncertainty}} in {{Artificial Intelligence}}},
  author = {Wimmer, Lisa and Sale, Yusuf and Hofman, Paul and Bischl, Bernd and H{\"u}llermeier, Eyke},
  editor = {Evans, Robin J. and Shpitser, Ilya},
  year = 2023,
  month = 8,
  series = {Proceedings of {{Machine Learning Research}}},
  volume = {216},
  pages = {2282--2292},
  publisher = {PMLR},
  address = {Pittsburgh, PA, USA},
  url = {https://proceedings.mlr.press/v216/wimmer23a.html}
}
\end{small}

\appendix
\section{LSTM Architecture and Training Details} \label{sec:lstm_model}
\subsection{Study Region}
This study focuses on the study region of western Canada, particularly the provinces of BC and Alberta. This region is characterized by large hydroclimatic variability \cite{schnorbusImpactsClimateChange2014} with a dense streamflow gauge network \cite{watersurveyofcanadaEnvironmentClimateChange2026} and well-documented occurrences of extreme events in recent years \cite{whiteUnprecedentedPacificNorthwest2023}.

\subsection{Data Sources and Preprocessing}

\begin{table}[H]
    \centering
    \caption{LSTM data used}
    \small
    \begin{tabular}{lll}
    \toprule
    \textbf{Handling} & \textbf{Variable name} & \textbf{Units}\\
    \midrule
    Dynamic input & Maximum temperature & $^\circ$C day$^{-1}$\\
    Dynamic input & Minimum temperature & $^\circ$C day$^{-1}$\\
    Dynamic input & Total precipitation & mm day$^{-1}$\\
    \midrule
    Static input & Catchment area & km$^2$\\
    Static input & Mean elevation & m above sea level\\
    Static input & Elevation range & m\\
    Static input & Mean slope & -\\
    Static input & Fractional glacier cover & -\\
    \midrule
    Output & Streamflow discharge & mm day$^{-1}$
    \end{tabular}
    
    \label{tab:datasets}
\end{table}

\paragraph{Streamflow Data:}
Daily mean streamflow target data (mm/day) and drainage basin polygons were obtained from the Environment and Climate Change Canada HYDAT database \cite{watersurveyofcanadaEnvironmentClimateChange2026, environmentandclimatechangecanadaNationalHydrometricNetwork2024}. The analysis was restricted to natural (unregulated) stream gauges located within the study region and active during the 1980--2022 period with no more than 50\% missing observations in any calendar year.

\paragraph{Meteorological Forcings:}
Hourly 2m air temperature and total precipitation were extracted from the ERA5-Land reanalysis dataset \cite{munozsabaterERA5LandHourlyData2019} at a resolution of $0.1^\circ\times0.1^\circ$ and spatially averaged over each catchment to derive daily precipitation, maximum temperature, and minimum temperature.

\paragraph{Static Catchment Attributes:}
Catchment topography was characterized using the Terrain Tiles digital elevation model \cite{mapzenTerrainTiles2026} to compute catchment area, mean elevation, elevation range, and mean slope over each catchment. Glacier extent was similarly derived using Version 6 of the Randolph Glacier Inventory \cite{rgiconsortiumRandolphGlacierInventory2017}.

\paragraph{Standardization:}
Prior to training, all predictor variables were standardized using the mean and standard deviation computed exclusively from the 1980–2005 training period to prevent data leakage.

\subsection{Dataset Partitions} 

\begin{table}[H]
    \centering
    \caption{Partitions of forcing and streamflow datasets. The forcing data subsets begin exactly one year prior to the streamflow subsets to provide the 365-day meteorological context window required for the first target prediction.}
    \small
    \begin{tabular}{lccc}
    \toprule
    \textbf{Subset} & \textbf{Forcing data} & \textbf{Streamflow data} & \textbf{Number of observations}\\
    \midrule
    Training & 1979/01/01--2005/12/30 & 1980/01/01--2005/12/31 & 2,278,178\\
    Validation & 2005/01/01--2012/12/30 & 2006/01/01--2012/12/31 & 607,967\\
    Testing & 2012/01/01--2022/12/30 & 2013/01/01--2022/12/31 & 877,729\\
    \bottomrule
    \end{tabular}
    \label{tab:dataset_partitions}
\end{table}

\subsection{Model Architecture}
The model consists of a single LSTM layer with a hidden-state dimension of 256, implemented in PyTorch. A fully connected output layer produces the final daily streamflow prediction for each input sequence. The network receives a sequence length of 365 days. Static catchment attributes are concatenated with the dynamic meteorological inputs at each timestep, allowing topographic characteristics to influence the evolving hidden state throughout the one-year context window. 

\subsection{Training Procedure} \label{sec:lstm_training}
\paragraph{Loss Function:}
The network was optimized using the basin-averaged Nash-Sutcliffe Efficiency ($\mathrm{NSE}^*$) loss function proposed by Kratzert et al. \cite{kratzertLearningUniversalRegional2019}:
\begin{equation}
\mathrm{NSE}^*=
\frac{1}{B}
\sum_{b=1}^{B}
\sum_{n=1}^{N_b}
\frac{(\hat{y}_{n,b}-y_{n,b})^2}
{(\sigma_b+\epsilon)^2},
\label{eq:nse_loss}
\end{equation}
where $B$ is the number of basins, $N_b$ is the number of observations for basin $b$, $\hat{y}_{n,b}$ and $y_{n,b}$ denote predicted and observed specific discharge, $\sigma_b$ is the standard deviation of observed discharge during the training period, and $\epsilon$ is a small constant introduced for numerical stability. This modified mean-squared-error objective normalizes residuals by the streamflow variability within each basin, preventing high-variance catchments from disproportionately influencing model training.

\paragraph{Masking:}
Missing streamflow observations within the dataset were masked during loss computation to avoid introducing interpolation errors.

\paragraph{Optimization:}
Model parameters were optimized using the Adam optimizer with an initial learning rate of 0.001, a batch size of 512, and a dropout rate of 0.4.

\paragraph{Ensemble Generation:}
Validation loss was evaluated after each epoch. Training was terminated using early stopping when the validation loss failed to improve for five consecutive epochs, up to a maximum of 50 epochs. This entire training procedure was independently repeated ten times using different random seeds to construct the $M=10$ random-initialization ensemble, with each member trained as an independent job. The random seeds govern both the initial network weights and the minibatch data ordering, providing the structural diversity between members.

\section{Estimator Definitions and Properties}
\subsection{Under-Dispersion Factor (UDF)} \label{sec:under-dispersion_factor}
The UDF is derived as the corrected inverse of the spread-skill ratio \cite{fortinWhyShouldEnsemble2014}.

Strata $\mathcal{S}$ are defined by within-catchment decile bounds of the
conditioning variable, either observed flow $y$ or the ensemble mean $\hat\mu$, pooled across catchments. The numerator of Eq.~(\ref{eq:under-dispersion}) is the standard deviation of residuals $r = y - \hat\mu$ within $\mathcal{S}$, computed with the unbiased $n-1$ convention after removing the pooled stratum mean; mean removal separates dispersion error from conditional bias, and the removed means are reported as conditional bias in Table~\ref{tab:ledger}.

The denominator is constructed as follows. For catchment-day $t$, let $f_{m,t}$ denote the prediction of member $m$ and 
\begin{equation}
    \hat\mu_t = \frac{1}{M}\sum_{m=1}^{M} f_{m,t}
\end{equation}
denotes the ensemble mean. The per-day between-member variance is computed with the unbiased convention,
\begin{equation}\label{eq:member_variance}
\hat\sigma^2_{e,t} = \frac{1}{M-1}\sum_{m=1}^{M}\left(f_{m,t}-\hat\mu_t\right)^2,
\end{equation}
and pooled over $\mathcal{S}$ by averaging variances before taking the root,
\begin{equation}\label{eq:pooled_variance}
\bar{\hat\sigma}^2_e = \frac{1}{|\mathcal{S}|}\sum_{t\in\mathcal{S}} \hat\sigma^2_{e,t}.
\end{equation}
The $(M-1)$ divisor in Eq.~(\ref{eq:member_variance}) makes $\hat\sigma^2_{e,t}$
an unbiased estimator of the member variance, which is required for the null of the corrected ratio to equal exactly $1$, the factor $\sqrt{(M+1)/M}$ corrects for the finite ensemble (Sect. \ref{sec:ensemble_correction}).

Pooling by averaging member standard deviations $\hat\sigma_{e,t}$ rather than variances $\hat\sigma^2_{e,t}$ understates spread: by Jensen's inequality, 
\begin{equation}
    \frac{1}
    {\vert\mathcal{S}\vert}\sum_{t\in\mathcal{S}} \sqrt{\hat\sigma^2_{e,t}} \le \sqrt{\frac{1}{|\mathcal{S}|}\sum_{t\in\mathcal{S}} \hat\sigma^2_{e,t}},
\end{equation}
with equality only when all per-day variances are equal, and they differ widely across a stratum. An understated spread overstates the factor. We pool variances throughout, the convention required for the null of Sect.~\ref{sec:ensemble_correction}, and the conservative one for our reported factors.

\subsection{Finite-Ensemble Correction and Exchangeability Null} \label{sec:ensemble_correction}
Ensemble members are trained as independent jobs differing in random initialization and minibatch data shuffling (Sect.~\ref{sec:lstm_training}), and are therefore exchangeable draws from the distribution over models induced by the shared data, architecture, and loss. Decomposing the error of member $m$ at input $x$ as
\begin{equation}
\varepsilon_m(x) = b(x) + \delta_m(x),
\label{eq:error_decomposition}
\end{equation}
where $b(x)$ is the systematic error of the training procedure, common to all members, and the $\delta_m$ are independent seed-driven deviations with variance $\sigma^2_\delta(x)$. Because $b(x)$ is shared, it cancels from every pairwise disagreement: the between-member variance (Eq.~(\ref{eq:member_variance})) is an unbiased estimator of $\sigma^2_\delta$ and is blind to $b$.

The null of Eq.~(\ref{eq:under-dispersion}) is that the observation is exchangeable with the ensemble members: writing $y = f^\star(x) + \varepsilon_0$, the truth behaves statistically as an $(M{+}1)$-th member. This holds exactly when $b(x) = 0$ and no aleatoric variability is left unmodeled, so that $\varepsilon_0$ is drawn from the same distribution as the $\delta_m$. Under this null, the residual to the ensemble mean is $r = \varepsilon_0 - \tfrac{1}{M}\sum_{m}\delta_m$, giving
\begin{equation}
\mathrm{Var}(r) = \sigma^2_\delta + \frac{\sigma^2_\delta}{M}
= \sigma^2_\delta\,\frac{M+1}{M},
\qquad
\mathbb{E}\!\left[\hat\sigma^2_{e,t}\right] = \sigma^2_\delta .
\label{eq:finite_correction}
\end{equation}
Dividing the reported spread by $\sqrt{(M+1)/M}$ therefore fixes the null of the corrected ratio at exactly $1$. A corrected factor above $1$ therefore reflects residual error components that are not represented by between-member disagreement, including components shared across members to which the spread is structurally blind. The same reliability condition, with the identical finite-$M$ factor, underlies spread-skill verification of ensemble forecasts in numerical weather prediction \cite{fortinWhyShouldEnsemble2014}.

\subsection{Conditioning on $\hat\mu$ and Interpretation of the Null} \label{sec:peak_bias}
Stratifying by observed flow selects on the outcome: a catchment-day enters the highest $y$-decile in part because its residual is positive, so under-predicted days are over-represented there, and the stratum mean residual is mechanically inflated. Conditioning on $\hat\mu$ uses only information the model itself produced, so the mean residual within a $\hat\mu$-stratum is genuine conditional bias; empirically, top-decile conditional bias falls from $+1.304$ to $+0.25$~mm\,day$^{-1}$ (Table~\ref{tab:ledger}). The per-day null of Sect.~\ref{sec:ensemble_correction} extends to strata directly: averaging $\mathbb{E}[r_t^2] = \sigma^2_{\delta,t}(M{+}1)/M$ over $\mathcal{S}$ preserves the ratio of the pooled quantities, provided variances are pooled before roots (Eq.~(\ref{eq:pooled_variance})).

The members are point predictors trained with a squared-error objective (Eq.~(\ref{eq:nse_loss})), so their disagreement reflects seed-driven variation only. Aleatoric variability enters the residual in the numerator, but not the spread, and raises the corrected ratio above $1$ even for an ideally dispersed ensemble. 

While observation error may grow with extreme flows due to the reliance on rating curve extrapolation, this observational uncertainty contributes to the residual dispersion and therefore provides one component of the observed UDF, but cannot by itself account for the rapid growth of the UDF. Literature assessing stage-discharge uncertainties indicates that typical error bounds for high flows are generally within 10\% to 20\% \cite{mansanarezRapidStageDischargeRating2019}, with extreme high-flow extrapolations notoriously difficult to gauge and carrying higher uncertainty \cite{morlotDynamicRatingCurve2014, mansanarezRapidStageDischargeRating2019}. The observed growth in the UDF cannot be interpreted purely as model-side error, but it demonstrates that the deployment-available spread becomes progressively less representative of realized uncertainty. Observation error contributes to residual dispersion and itself varies with flow, so the absolute magnitude of the factor should be interpreted cautiously. The diagnostic quantity throughout is accordingly the growth of the factor, rather than its absolute magnitude.

We note that this argument requires no separation of aleatoric from epistemic uncertainty, a decomposition whose identifiability for deep models is itself contested \cite{hullermeierAleatoricEpistemicUncertainty2021, wimmerQuantifyingAleatoricEpistemic31}, since both the realized dispersion and the reported spread are directly measurable. The aleatoric uncertainty would also be unavailable in operation unless uncertainty-aware learning or another capturing technique is implemented.

\subsection{Superlinear Scaling of Error with Spread} \label{sec:recalibration}
The scaling relation in Fig.~\ref{fig:dispersion}b is fit by ordinary least squares, regressing $\log \mathrm{SD}(r \mid \mathcal{S})$ against log-corrected pooled spread across the nine lowest $\hat\mu$-deciles, with the top decile held out as an out-of-sample check. A slope of $1$ would indicate proportional scaling between residual dispersion and spread; the fitted slope exceeds it ($\beta = 1.30$, $95\%$ CI $[1.20, 1.41]$), with intervals obtained by resampling catchments with replacement ($B{=}1000$, percentile intervals), recomputing strata and the regression within each replicate.

Superlinearity carries a direct consequence for post-hoc recalibration: for any constant rescaling $c > 0$ of the spread, the calibration ratio of the rescaled band, $\mathrm{SD}(r \mid s)/(c\,s) \propto s^{\beta-1}$, remains increasing in $s$, so no single $c$ calibrates the band across the fitted range. Closing the gap, therefore, requires calibration conditioned on flow.

\section{Evaluation Metrics}

\subsection{Nash-Sutcliffe Efficiency (NSE)}
Model predictive skill was evaluated using the standard NSE \cite{nashRiverFlowForecasting1970}. NSE compares the variance of the model residuals to the variance of the observed data:
\begin{equation}
\mathrm{NSE} = 1 - \frac{\sum_{t=1}^{T} (y_t - \hat{y}_t)^2}{\sum_{t=1}^{T} (y_t - \bar{y})^2},
\label{eq:nse}
\end{equation}
where $T$ is the total number of timesteps, $y_t$ is the observed streamflow at time $t$, $\hat{y}_t$ is the predicted streamflow at time $t$, and $\bar{y}$ is the mean of the observed streamflow over the evaluation period. An NSE of $1.0$ indicates perfect predictive agreement, while an NSE of $0$ indicates the model is only as accurate as predicting the long-term observational mean.

\subsection{Root Mean Squared Error (RMSE)} \label{sec:RMSE}
The absolute magnitude of prediction errors was quantified using the Root Mean Squared Error (RMSE). Unlike NSE, which provides a normalized relative measure, RMSE is expressed in the same physical units as the observed streamflow and disproportionately penalizes larger errors due to the squared term:
\begin{equation}
\mathrm{RMSE} = \sqrt{ \frac{1}{T} \sum_{t=1}^{T} (y_t - \hat{y}_t)^2 }.
\label{eq:rmse}
\end{equation}
$y_t$ and $\hat{y}_t$ are the observed and predicted streamflows at time $t$, and $T$ is the total number of timesteps. An RMSE of $0$ indicates perfect predictive accuracy, with larger values reflecting a greater average magnitude of deviation from the observations.

\subsection{Area Under Sparsification Error (AUSE)} \label{sec:AUSE}
The ranking performance of the ensembles' uncertainty signal was evaluated using the Area Under Sparsification Error (AUSE) \cite{ilgUncertaintyEstimatesMultihypotheses2018}. AUSE assesses how well the predicted uncertainties rank the actual prediction errors. We define a sparsification error curve, $E_s(\phi)$, as the error (e.g., RMSE) calculated on the remaining fraction of data $\phi \in [0,1]$ after masking out the $1-\phi$ fraction of points with the highest predicted variance. This is compared to an oracle curve, $E_o(\phi)$, which calculates the error after masking out the $1-\phi$ fraction of points with the highest actual squared errors. The AUSE is the integrated area between these two curves:
\begin{equation}
\mathrm{AUSE} = \int_{0}^{1} \left( E_s(\phi) - E_o(\phi) \right) d\phi.
\label{eq:ause}
\end{equation}
An AUSE of $0$ indicates perfect uncertainty ordering, meaning the model's highest predicted uncertainties perfectly correspond to its largest actual errors. Larger AUSE values indicate poorer ordering of the predictive uncertainty.

\subsection{Spearman's Rho} \label{sec:spearmans-rho}
To assess the monotonic relationship between the per-catchment AUSE and the UDF without assuming linearity, Spearman's rank correlation coefficient ($\rho$) was utilized. This non-parametric metric relies on the rank order of the metrics across catchments rather than their absolute magnitudes:
\begin{equation}
\rho = 1 - \frac{6 \sum_{i=1}^{N} d_i^2}{N(N^2 - 1)},
\label{eq:spearman}
\end{equation}
where $d_i = \mathrm{rank}(\mathrm{AUSE}_i) - \mathrm{rank}(\text{UDF}_i)$ represents the difference in the ordinal rank of the AUSE and the UDF for the $i$-th catchment, and $N$ is the total number of catchments. Spearman's $\rho$ ranges from $-1.0$ to $1.0$. A value of $1.0$ indicates a perfect monotonically increasing relationship, $0$ indicates no monotonic correlation, and $-1.0$ indicates a perfect monotonically decreasing relationship.

\subsection{Log-log slope for Simple Linear Regression} \label{sec:OLS}
To assess whether the ensemble spread scales proportionally with realized error, a simple linear regression was fitted to the log-transformed variables across deciles 10 through 90. Let $s_i$ be the reported ensemble spread and $\sigma_i$ be the residual standard deviation for the $i$-th decile. The relationship is modeled as:
\begin{equation}
\log(\sigma_i) = \beta_0 + \beta_1 \log(s_i) + \epsilon_i,
\label{eq:loglog}
\end{equation}
where $\beta_1$ represents the log-log slope. The slope parameter $\beta_1$ is estimated using Ordinary Least Squares (OLS):
\begin{equation}
\hat{\beta}_1 = \frac{\sum_{i=1}^{n} (\log(s_i) - \overline{\log(s)})(\log(\sigma_i) - \overline{\log(\sigma)})}{\sum_{i=1}^{n} (\log(s_i) - \overline{\log(s)})^2}.
\label{eq:ols_slope}
\end{equation}
A slope of $\hat{\beta}_1 = 1$ indicates proportional scaling, whereas $\hat{\beta}_1 > 1$ indicates that residual dispersion grows faster than the reported ensemble spread.

\section{Data Tables}

\subsection{Figure 1} \label{sec:figure1_datatable}
\begin{table}[H]
    \centering
    \small
    \setlength{\tabcolsep}{5pt}
    \caption{The Conditional Bias, UDF, and mean residual $r=y-\hat\mu$ behind Figure~\ref{fig:dispersion}, reported by decile of each conditioning variable. Deciles are within-catchment flow percentiles; ties in observed discharge leave the $y$-conditioned deciles slightly uneven. The null for the UDF is exactly~1 under $\hat\mu$-conditioning; the null for the mean residual is~$0$. Top-decile catchment-clustered bootstrap 95\% intervals ($B=1000$): $5.76$~[$4.48$,~$6.90$] conditioned on $y$, $4.27$~[$3.50$,~$4.99$] conditioned on $\hat\mu$. Bold outlines the out of sample test.}
    \label{tab:ledger}
    \begin{tabular}{lrcrrcr}
    \toprule
    & \multicolumn{3}{c}{conditioned on $y$} & \multicolumn{3}{c}{conditioned on $\hat\mu$} \\
    \cmidrule(lr){2-4}\cmidrule(lr){5-7}
    decile & $n$ & UDF & mean $r$ & $n$ & UDF & mean $r$ \\
    \midrule
    0--10 & 82{,}317 & 2.00 & $-0.096$ & 87{,}680 & 2.10 & $-0.010$ \\
    10--20 & 88{,}420 & 1.97 & $-0.086$ & 87{,}735 & 2.85 & $-0.018$ \\
    20--30 & 90{,}156 & 2.16 & $-0.101$ & 87{,}750 & 2.53 & $-0.023$ \\
    30--40 & 96{,}425 & 2.16 & $-0.117$ & 87{,}735 & 2.65 & $-0.028$ \\
    40--50 & 84{,}593 & 2.23 & $-0.160$ & 87{,}727 & 3.26 & $-0.031$ \\
    50--60 & 85{,}847 & 2.16 & $-0.172$ & 87{,}910 & 3.52 & $-0.031$ \\
    60--70 & 86{,}898 & 2.20 & $-0.174$ & 87{,}746 & 4.00 & $-0.007$ \\
    70--80 & 87{,}409 & 2.34 & $-0.168$ & 87{,}739 & 4.05 & $-0.023$ \\
    80--90 & 87{,}734 & 2.27 & $-0.083$ & 87{,}746 & 4.37 & $+0.072$ \\
    \bfseries 90--100 & \bfseries 87{,}930 & \bfseries 5.76 & \bfseries $+1.304$ & \bfseries 87{,}961 & \bfseries 4.27 & \bfseries $+0.250$ \\
    \bottomrule
    \end{tabular}
\end{table}

\subsection{Figure 2} \label{sec:figure2_datatable}
\begin{table}[H]
    \centering
    \small
    \setlength{\tabcolsep}{5pt}
    \caption{Sparsification curves behind Figure~\ref{fig:shift}a, on the plotted grid of $91$ points, with removal in $1\%$ steps to $90\%$ removed, sampled here every tenth point. Each curve is normalized by its set's full-sample RMSE, so every ordering starts at $1.000$ and AUSE is unaffected by the normalization. Spread orders catchment-days by reported ensemble variance, flow by the ensemble mean prediction $\hat\mu$ alone, oracle by realized squared error, and random by a seeded permutation averaged over ten draws, flat to within $0.016$. The flow column is the comparator for the ordering claim; it carries no uncertainty information, only predicted magnitude, and is computable at deployment. These are integrated over a $91$-point grid, not over the ten rows shown.}
    \begin{tabular}{lrrrrcrrrr}
    \toprule
    & \multicolumn{4}{c}{all test} & & \multicolumn{4}{c}{top flow decile} \\
    \cmidrule(lr){2-5}\cmidrule(lr){7-10}
    fraction removed & spread & flow & oracle & random & & spread & flow & oracle & random \\
    \midrule
    $0.0$ & $1.000$ & $1.000$ & $1.000$ & $1.000$ & & $1.000$ & $1.000$ & $1.000$ & $1.000$ \\
    $0.1$ & $0.294$ & $0.351$ & $0.139$ & $1.000$ & & $0.610$ & $0.670$ & $0.208$ & $1.002$ \\
    $0.2$ & $0.183$ & $0.217$ & $0.080$ & $1.003$ & & $0.465$ & $0.575$ & $0.144$ & $0.997$ \\
    $0.3$ & $0.129$ & $0.157$ & $0.052$ & $1.002$ & & $0.389$ & $0.483$ & $0.107$ & $0.998$ \\
    $0.4$ & $0.097$ & $0.117$ & $0.036$ & $0.999$ & & $0.302$ & $0.422$ & $0.081$ & $0.997$ \\
    $0.5$ & $0.075$ & $0.085$ & $0.025$ & $1.002$ & & $0.279$ & $0.347$ & $0.062$ & $0.997$ \\
    $0.6$ & $0.059$ & $0.069$ & $0.017$ & $1.005$ & & $0.223$ & $0.293$ & $0.045$ & $0.998$ \\
    $0.7$ & $0.045$ & $0.054$ & $0.011$ & $1.016$ & & $0.199$ & $0.233$ & $0.032$ & $0.998$ \\
    $0.8$ & $0.032$ & $0.037$ & $0.007$ & $1.016$ & & $0.155$ & $0.216$ & $0.020$ & $0.987$ \\
    $0.9$ & $0.020$ & $0.020$ & $0.003$ & $1.016$ & & $0.126$ & $0.138$ & $0.009$ & $1.004$ \\
    \bottomrule
    \end{tabular}
\end{table}

\begin{table}[H]
    \centering
    \small
    \setlength{\tabcolsep}{5pt}
    \caption{The 269 catchments of Figure \ref{fig:shift}b, binned into deciles of per-catchment AUSE. Per-catchment rows are omitted for space, so the panel's claim is given in its smallest form: sorting catchments by how well their spread ranks error does not sort them by how well it is scaled. UDF is computed within each catchment; Spearman $\rho = -0.23$ ($p = 0.00014$). Every catchment $(n=269)$ exceeds the null of~$1$, minimum $1.03$.}
    \label{tab:perbasin_data}
    \begin{tabular}{lrcrrr}
    \toprule
    & & & \multicolumn{3}{c}{UDF} \\
    \cmidrule(lr){4-6}
    AUSE decile & $n$ & AUSE range & median & min & max \\
    \midrule
    0--10 & $27$ & $[0.024, 0.071]$ & $3.45$ & $1.68$ & $10.40$ \\
    10--20 & $27$ & $[0.072, 0.092]$ & $3.14$ & $1.17$ & $5.94$ \\
    20--30 & $27$ & $[0.093, 0.105]$ & $2.44$ & $1.38$ & $5.25$ \\
    30--40 & $27$ & $[0.106, 0.122]$ & $2.16$ & $1.39$ & $4.68$ \\
    40--50 & $26$ & $[0.123, 0.136]$ & $2.44$ & $1.03$ & $10.78$ \\
    50--60 & $27$ & $[0.137, 0.155]$ & $1.96$ & $1.27$ & $8.32$ \\
    60--70 & $27$ & $[0.156, 0.172]$ & $2.37$ & $1.34$ & $6.24$ \\
    70--80 & $27$ & $[0.173, 0.192]$ & $2.02$ & $1.16$ & $7.36$ \\
    80--90 & $27$ & $[0.192, 0.230]$ & $2.13$ & $1.32$ & $6.57$ \\
    90--100 & $27$ & $[0.231, 0.470]$ & $2.48$ & $1.14$ & $8.25$ \\
    \bottomrule
    \end{tabular}
\end{table}

\section{Code and Data Availability}
All data and code used to conduct this study are publicly available. ERA5-Land data were accessed from \cite{copernicusclimatechangeservicec3sERA5LandHourlyData2019}. Streamflow data and catchment polygons were accessed from \cite{environmentandclimatechangecanadaNationalHydrometricNetwork2024, watersurveyofcanadaEnvironmentClimateChange2026}. DEM data were accessed from \cite{mapzenTerrainTiles2026}. Glacier outlines were accessed from \cite{rgiconsortiumRandolphGlacierInventory2017}. All code and preprocessed data used to conduct the analysis and reproduce the figures and tables created in this study are available from \cite{gallantLSTMEnsembleUnderdispersion2026}.

\end{document}